\documentclass[conference]{IEEEtran}
\usepackage{booktabs}
\IEEEoverridecommandlockouts

\def\bng{\bngx}

\font\bngx=bang10

\def\*#1*#2{o\null{#2}{#1}}

\def\sh#1{\setbox0=\hbox{#1}%
     \kern-.02em\copy0\kern-\wd0
     \kern.04em\copy0\kern-\wd0
     \kern-.02em\raise.0433em\box0 }
\usepackage{cite}
\usepackage{amsmath,amssymb,amsfonts}
\usepackage{hyperref}
\usepackage{xcolor}
\usepackage{multirow}
\usepackage{url}
\usepackage{subcaption}
\usepackage{float}
\usepackage{booktabs,chemformula}
\usepackage{hyperref}
\usepackage{caption}
\usepackage{cleveref}
\usepackage{fancyhdr}
\usepackage[super]{nth}
\usepackage[pscoord]{eso-pic}
\usepackage{amssymb}
\usepackage{float}
\usepackage{tikz}
\usepackage{textcomp}
\usepackage{lipsum}
\hypersetup{
    colorlinks,
    linkcolor=blue,
    citecolor=blue,
    urlcolor=blue
}

\usepackage{algorithmic}
\usepackage{graphicx}
\usepackage{textcomp}
\usepackage{xcolor}
\usepackage{caption}
\usepackage{siunitx} 
\def\BibTeX{{\rm B\kern-.05em{\sc i\kern-.025em b}\kern-.08em
    T\kern-.1667em\lower.7ex\hbox{E}\kern-.125emX}}

\newcommand\copyrighttext{%
  \footnotesize \textcopyright 2023 IEEE. Personal use of this material is permitted.
  Permission from IEEE must be obtained for all other uses, in any current or future
  media, including reprinting/republishing this material for advertising or promotional
  purposes, creating new collective works, for resale or redistribution to servers or lists, or reuse of any copyrighted component of this work in other works.}
\newcommand\copyrightnotice{%
\begin{tikzpicture}[remember picture,overlay]
\node[anchor=south,yshift=10pt] at (current page.south) {\fbox{\parbox{\dimexpr\textwidth-\fboxsep-\fboxrule\relax}{\copyrighttext}}};
\end{tikzpicture}%
}

\makeatletter

\let\old@ps@IEEEtitlepagestyle\ps@IEEEtitlepagestyle
\def\confheader#1{%
    \def\ps@IEEEtitlepagestyle{%
        \old@ps@IEEEtitlepagestyle%
        \def\@oddhead{\strut\hfill#1\hfill\strut}%
        \def\@evenhead{\strut\hfill#1\hfill\strut}%
    }%
    \ps@headings%
}
\makeatother

\confheader{%
    \parbox{50cm}{\textbf{Accepted and presented at "2023 26th International Conference on\\ Computer and Information Technology (ICCIT) 13-15 December 2023, Cox’s Bazar, Bangladesh"}}
}

\begin{document}

\title{Bengali License Plate Recognition: Unveiling Clarity with CNN and GFP-GAN}


\author{\IEEEauthorblockN{Noushin Afrin, Md Mahamudul Hasan, Mohammed Fazlay Elahi Safin, 
  Khondakar Rifat Amin,\\Md Zahidul Haque, Farzad Ahmed, and Md. Tanvir Rouf Shawon}
\IEEEauthorblockA{\textit{Department of Computer Science and Engineering,} \\
\textit{Ahsanullah University of Science and Technology, Dhaka, Bangladesh}\\
noushinmaifa@gmail.com, mhtusar120725@gmail.com, fazlayelahisafin@gmail.com, khrifatamin@gmail.com,\\
  jahidprototype2@gmail.com, farzad.cse@aust.edu, shawontanvir95@gmail.com}
}

\maketitle
\copyrightnotice

\begin{abstract}
Automated License Plate Recognition(ALPR) is a system that automatically reads and extracts data from vehicle license plates using image processing and computer vision techniques. The Goal of LPR is to identify and read the license plate number accurately and quickly, even under challenging, conditions such as poor lighting, angled or obscured plates, and different plate fonts and layouts. The proposed method consists of processing the Bengali low-resolution blurred license plates and identifying the plate's characters. The processes include image restoration using GFPGAN, Maximizing contrast, Morphological image processing like dilation, feature extraction and Using Convolutional Neural Networks (CNN), character segmentation and recognition are accomplished. A dataset of 1292 images of Bengali digits and characters was prepared for this project. 
\end{abstract}


\section{\textbf{INTRODUCTION}}

Automated License Plate Recognition (ALPR) technology has revolutionized the field of vehicle identification by automating the extraction of information from license plates. Through cutting-edge computer vision techniques, ALPR systems capture license plate photos, segment characters, and accurately identify them. In our study, we place primary emphasis on developing an innovative ALPR system that harnesses the power of deep learning and image processing techniques, specifically optimized for Bengali license plates. Number plates consist of digits, registered letters and city names. The digits serve as unique identifiers for each vehicle, while the registered letters indicate the vehicle type. The city name on the license plate provides information about the vehicle's registration location. Accurately identifying and interpreting these components allows us to extract specific details and understand the vehicle's identification, type, and origin. 

However, identifying license plates from poor-quality images presents significant challenges. Issues such as poor lighting, diverse plate fonts, and variations in sizes and orientations complicate the precise segmentation and recognition of characters. 

Bengali symbols and numerals are a little more complicated than English characters and digits, making it difficult to recognize number plates in Bengali. There have been prior studies on Bengali license plate recognition \cite{r1,r2} and digit recognition \cite{10103341,khan2023evaluation}, but none have taken into account the low image quality that is prevalent for street cameras, which are the primary source of license plate images.\\

Our research's main goal is to build advanced methods that can accurately recognize and evaluate the components of Bengali license plates, even in low-quality photographs. In addition, we utilize GFPGAN for image enhancement and leverage the state-of-the-art CNN and ResNet-50 models for the task of recognizing the digits and characters of the number plates. By showcasing the capabilities of deep learning and image processing, our research aims to pave the way for improved ALPR systems. These advancements will have widespread applications, benefiting numerous industries that rely on accurate vehicle identification. The results of our project will contribute to the development of ALPR systems with enhanced efficiency and precision, enabling more reliable and effective vehicle identification. The contribution of the paper can be summarized into the following points:


\begin{itemize}
  
  \item Modern deep learning methods are used by us to perform  Bengali license plate recognition in challenging environments, including Convolutional Neural Networks (CNNs) and pre-trained models like ResNet-50.

  \item We incorporate GFPGAN for image enhancement, further improving the quality of input images and enhancing recognition performance.

  \item A dataset of 1292 images of Bengali digits and characters was prepared for this project.
  
  \item We scored an excellent 88\% average accuracy in accurately identifying the 30 various number plates tested by the SequenceMatcher class of difflib\footnote{\url{https://docs.python.org/3/library/difflib.html}} library. The codes and dataset can be found at - \url{https://github.com/mhtusher131/Bengali-License-Plate-Recognition_Unveling-Clarity-with-CNN-and-GFP-GAN?fbclid=IwAR0PLgVXkW3lJ8huKg_5vqDtK4qPOMv2u-I-iegGoibDJzvhNpxpPjTObJ4}

\end{itemize}

\section{\textbf{LITERATURE REVIEW}}

This literature review provides an overview of the advancements made in detecting number plates in Bengali as well as in other languages from poor-quality images.

\textbf{Study on License Plate Recognition Using CNNs:}
These three papers \cite{r3, r4, r5} focus on license plate recognition systems that use convolutional neural networks (CNNs) and other computer vision techniques. Munna et al.\cite{r3} describes a systemfor utilizing a CNN to find and identify license plate, with a specific focus on Bengali characters. Suvon et al.\cite{r4} propose a real-time system for recognizing Bengali license plates using a combination of computer vision techniques and a CNN. Dhar et al.\cite{r5} present a method that uses edges to recognize license plates detection and a CNN for feature extraction and character recognition. These papers collectively demonstrate the effectiveness of using CNNs for license plate recognition, which has important applications in traffic control, surveillance, and tracking of stolen vehicles.

\textbf{Study on Automatic License Plate Recognition for Vehicle Authentication in Bengali:} The study by Nooruddin et al.\cite{r1} and Islam et al.\cite{r2}, focuses on Bengali Automatic License Plate Detection and Recognition for Vehicle Authentication. The paper\cite{r1} by Nooruddin et al. uses a Random Forest model for the area-based detection and identification of license plates aspect ratio criteria. The paper \cite{r2} by Islam et al. introduces a unique four-step system including preprocessing, license plate processing, character recognition, and vehicle registration determination. This approach achieves a 96.1\% character recognition score using techniques like character segmentation, edge detection, and template matching.

\textbf{Study on Automatic License Plate Recognition for Vehicle Authentication in other languages:} This study \cite{kathirvel2023systematic} by Kathirvel et al. presents a systematic number plate recognition (SNPR) methodology using YOLOv5s for automating the fine collection and enforcing speed limits on Indian roads. The proposed system achieves 98.2\% accuracy on number plate identification task. The study\cite{antar2022automatic} by Antar R et al. explores automatic license plate recognition for Saudi car plates. The experiment used 50 images, employing canny edge detection, threshold techniques, and horizontal projection for plate segmentation. OCR was applied to read English and Arabic characters separately,obtaining an accuracy of 96\% for English and 92.4\% for Arabic. Raj R et al.\cite{raj2022license} implements an ANPR system using YOLOv5 for number plate detection. 
The paper \cite{kabiraj2023number} by kabiraj et al. [2023] offers a method for enhancing number plate detection and identification accuracy using Enhanced-Super-Resolution with Generative Adversarial Network (ESRGAN).


\section{\textbf{BACKGROUND STUDIES}}

\textbf{GFPGAN:} Generative Flow-Based Probabilistic Graphical Model\cite{yin2022styleheat} is a type of generative model that uses a probabilistic graphical model to generate data from a given set of input variables. It is based on the idea of using flow-based generative models to learn the underlying structure of data and then generate new data from it. GFPGAN has been used in various applications such as image generation, text generation, and music generation. It has also been used for anomaly detection and forecasting tasks. GFPGAN is an efficient and powerful tool for generating high-quality data from complex distributions.

\textbf{ResNet-50:} The motivation for developing deep neural networks stemmed from challenges in training deep networks, including the degradation of performance and vanishing gradients. ResNet, a significant architecture, emerged to tackle these issues, with ResNet-50\cite{patil2023deep} as a specific variant featuring 50  layers and shortcut connections that enable direct information flow, addressing vanishing gradients and aiding deep network training while using residual blocks to capture intricate features effectively.

\textbf{Performance Metrics:} 
Recall is a measurement of the proportion of positive cases that the classifier properly predicted out of all the positive cases in the data, whereas precision is measured by the percentage of correctly predicted positive outcomes (TP). The harmonic mean of recall and precision is F1.

\section{\textbf{DATASET}}
We have created the dataset for Bengali License Plate Detection from Poor Quality Images. The dataset creation process involved several steps to collect and organize the Bengali number plate images:

\begin{figure}[!h]

  \centering
  \includegraphics[width=0.5\textwidth]{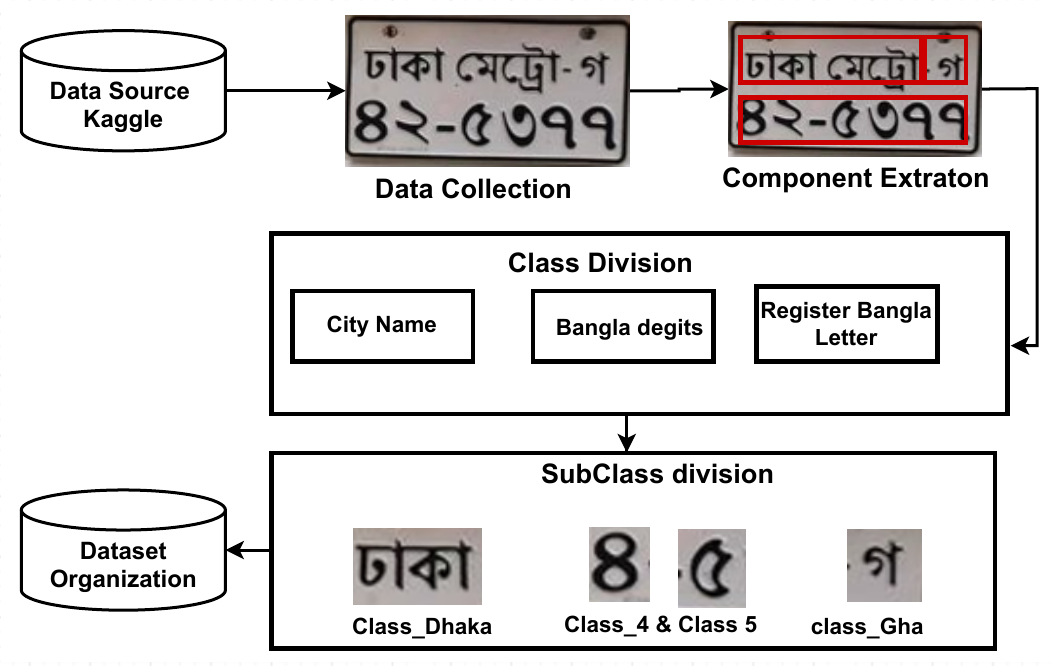}
   \caption{ Navigating Dataset Development Pipelines- Strategies and Insights}
\end{figure}

{\textbf{Data Collection:} A diverse collection of Bengali number plate images was gathered from various sources\footnote{\url{https://www.kaggle.com/furcifer/bangla-license-plates-synthetic}}. The images were collected to encompass a wide range of vehicle types, number plate designs, and environmental conditions\footnote{\url{https://github.com/mmshaifur/BLRPS}}. 

{\textbf{Component Extraction:} From each collected image, the specific components of interest were extracted. These components included digits, registered letters, city names, and metro names. Manual image cropping techniques were employed to precisely isolate these components, removing any unnecessary background information.



{\textbf{Class Division:} The extracted components were categorized into three main classes: digits, registered letters, and city names. This allowed for separate analysis and classification of each component type.

{\textbf{Subclass Division:} To capture variations within each class, subclass division was implemented. This resulted in 10 subclasses for digits, 4 subclasses for registered letters, and 3 subclasses for city names. The subclasses were designed to represent specific patterns and variations found in the dataset.

{\textbf{Dataset Organization:} The final dataset was organized into subclasses. The images were standardized to a size of 32x32 pixels with RGB color channels, ensuring consistency across the dataset.


\subsection{Dataset Statistics}
Table \ref{tab:statistical-table} presents the statistical information of the dataset.
The dataset consists of three main classes: digits, registered letters, and city names. Each class is further divided into subclasses, as specified. The number of images per subclass is satisfactory, ensuring sufficient data for model training and evaluation. In our dataset the city names are 'Dhaka ({\bng Dhaka})', 'Dhaka Metro ({\bng Dhaka emeTRa})', 'Narayanganj ({\bng narayNgNJ/j})'. The registered letters are Ka ({\bng k}), ga ({\bng g}), Jha ({\bng jh}), and La ({\bng l}).The digits are from 0 ({\bng 0})  to 9 ({\bng 9}).
 Bangladesh employs a diverse range of cities and registered letters on its number plates, reflecting the extensive coverage of these elements in its transportation system. These selections serve as the foundation for constructing a model that recognizes Bangladeshi number plates in low-quality images. Furthermore, the dataset's flexibility allows for potential expansions of additional cities and letters as required for the project's goals. Fig. \ref{fig:samples} depicts some samples from our dataset.


\begin{table}
   \centering
  \caption{Class distribution among train, validation and test set.}
    \begin{tabular}{ccccc}
    \hline
    \textbf{Class}& \textbf{Subclass type} & \textbf{Train} & \begin{tabular}[c]{@{}c@{}}valid-\\ation\end{tabular}& \textbf{Test} \\
    \hline
    & 0 ({\bng 0}) & 90 & 35 & 22    \\
    & 1 ({\bng 1}) & 140 & 35 & 27   \\
    & 2 ({\bng 2}) & 130  & 35 & 23   \\
     & 3 ({\bng 3}) & 130 & 35 & 22 \\
     & 4 ({\bng 4}) & 100  & 35 & 23\\

    Digits  & 5 ({\bng 5}) & 140 & 35 & 24 \\
     & 6 ({\bng 6}) & 90 & 35 & 26  \\
     & 7 ({\bng 7}) & 120 & 35 & 20  \\
     & 8 ({\bng 8}) & 110  & 35 & 21  \\
    & 9 ({\bng 9}) & 1 & 45 & 21 \\
    \hline
      & Ka ({\bng k})  & 13  & 7 & 7 \\
      & ga ({\bng g}) & 15 & 9 & 9\\
     registered 
      & jha ({\bng jh}) & 30   & 13 & 14\\
      & La ({\bng l}) & 18  & 12 & 12 \\
    \hline
     & Dhaka ({\bng Dhaka}) & 18  & 11 & 12\\
    City & Narayanganj ({\bng narayNgNJ/j}) & 30  & 15 & 10\\
     & Metro ({\bng emeTRa}) & 18   & 12 & 10\\
    \hline
    \end{tabular}
    \label{tab:statistical-table}
\end{table}

\begin{figure}[!h]
  \centering
  \includegraphics[width=0.5\textwidth]{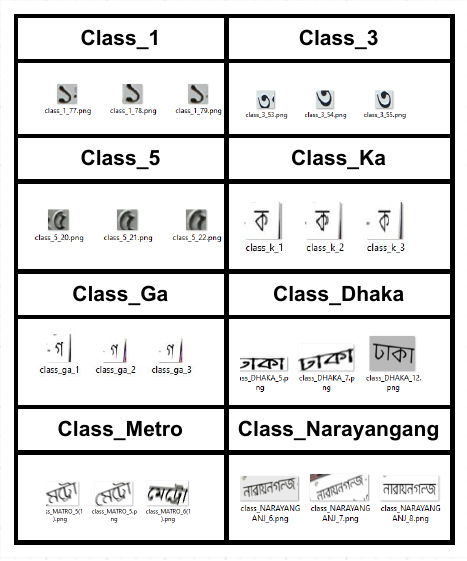}
   \caption{Illustration of the samples from our dataset.}
   \label{fig:samples}
\end{figure}




%







 \section{\textbf{METHODOLOGY}}

This section describes our methodology in brief. Fig. \ref{fig:img1} shows the workflow of our methodology. The ten individual steps of our methodology are described elaborately.\\






 
\textbf{1) Read the image:} This is the initial phase of our suggested procedure. Python at this point captures and reads the car picture. The image is blurry and of poor quality, making it difficult to read without altering how the read image or test image is depicted.



\textbf{2) Resize Image:} We must resize the image to a predetermined size because the acquired image may be of any size.Using the provided method, the height of the car image is converted into 32*32 pixels.



  \textbf{3) GFPGAN:} GFPGAN \cite{yin2022styleheat}(Generative Focused Pixel-wise GAN). It is a type of generative model that can be used to generate new images from existing ones. GFPGAN uses a combination of convolutional neural networks (CNNs) and Using generative adversarial networks (GANs), high-quality images can be produced from inputs with low resolution..GFPGAN has been used in low-resolution blur images to increase the quality of the image by reducing the blurriness.




  \textbf{4) RGB to greyscale picture conversion:}  To be more precise,The image's red, green, and blue parts are separated and then combined to form a single grey image.
  
  \textbf{5) Grey-scale picture binarization:} Using the Otsu method, the grey image is next transformed into a binary image. In this method, the conversion procedure is started after determining a threshold value.
  


\textbf{6) Maximising Contrast:} Maximising the contrast of an image is the process of increasing the difference between the lightest and darkest areas of an image. It is done by adjusting the brightness, contrast, and gamma levels of the image.






\textbf{7) Morphological Operation:} Dilation has been used to fill up the noise within the objects of the image.

\textbf{8) Extracting the License Plate Region:} In the suggested procedure, one of the key steps is the extraction of the license plate region.
a) Multiplication of the images: To get an output that only contains the license plate region, we now multiply the dilated image by the grayscale image.
b) Match contours: A technique for identifying the edges of numbers and letters in an image is called match contours. In order to do this, a set of predetermined shapes is matched up with the edges of the image's objects. This method is being used to classify, identify forms, and identify the items. The image's potential dimensions are obtained, the binary image's counter is found, and the coordinates of the enclosed rectangle are then provided. The contour's dimensions are then examined in order to separate the characters based on their size. After extracting each character using the coordinates of the surrounding rectangle, the output was prepared for categorization.



\textbf{9) Character Segmentation:} A license plate from Bangladesh contains two lines of text. The first line consists of a Bengali letter and one or two Bengali words. The second line contains six numbers as required by the BRTA's Retro-Reflective standard. Separations exist between the six digits. However, "Matra" is used to connect the alphabetic characters in a word (a line joins several characters). It is believed that the entire word "Matra" consists of one connected component. Here, linked component analysis is utilized to segment the characters. Following the binarization of the recovered license plate image, blue boundary boxes are created for each of the associated items in the shown image. After that, the characters are taken out of the plate and stored as a separate image. The recognition module receives the recently segmented characters and words.

\textbf{10) Recognition of the characters:} At this point, it should be possible to identify the characters that have already been split and saved as separate images. The CNN model is employed in the suggested method. The CNN model is trained on a sizable dataset of tagged images before being used to recognize segmented images utilizing datasets. The model was then able to learn the characteristics of each image and categorize them. With 1292 photos of Bengali characters and numbers, we trained the model.  The model can then be used to categorize fresh photos after being trained. Images that have been segmented have been highly correctly sorted into their respective groups using CNN models.

\begin{figure}[!h]
    \centering
    \includegraphics[width=0.53\textwidth]{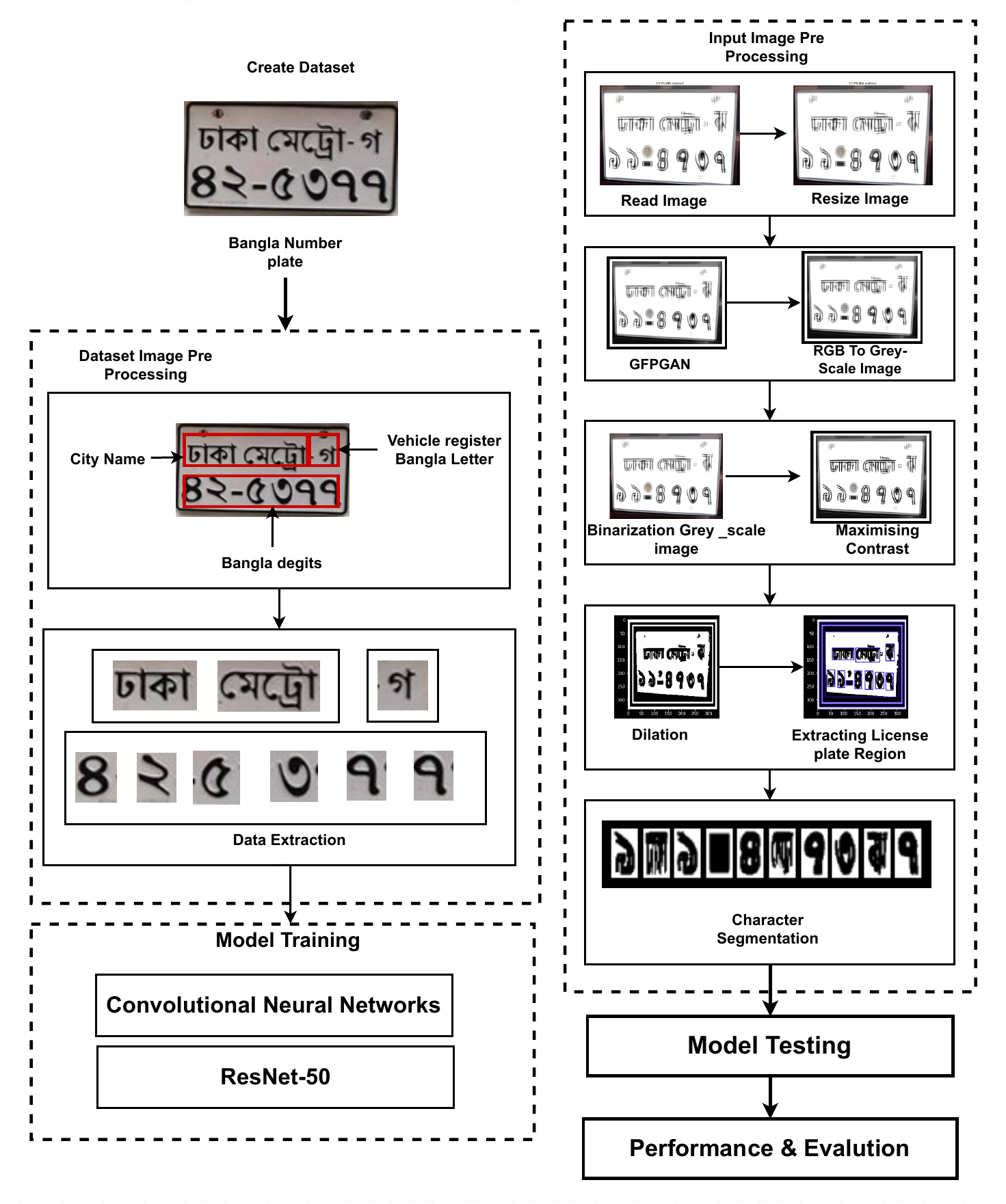}
    \caption{A visual guide to the proposed approach. }
    \label{fig:img1}
\end{figure}
\section{ \textbf{RESULTS AND EXPERIMENTS} }
\subsection{ \textbf{Hyperparameter Settings} }
In our instance, we used images with a size of $32,32,3$ to train the CNN and ResNet50 models. In both models 1292 images were used for training, 429 images were used for validation and 303 images were used to conduct tests.
The learning rate was 0.0001 and the entire batch length as 1.

\subsection{  \textbf{Performance of the Proposed Models} }
 The  general effectiveness of CNN and ResNet50 is given in Table \ref{tab:comp}. We presented the confusion matrix of our best-performing CNN architecture in Fig. \ref{fig:img2}. 

 \begin{table}[h!]
      \centering
      \caption{ Performance of Different Architectures (CNN and ResNet50) on the Test Dataset}
      \label{tab:comp}
\begin{tabular}{ c c c c c }

 \hline
\textbf{Architecture}  & \textbf{Accuracy}  & \textbf{Precision} & \textbf{Recall} & \textbf{F1-score} \\
 \hline
CNN   & 92.38 & 0.9667 & 0.9162 & 0.9476  \\

ResNet50   & 87.50 & 0.9343 & 0.8827 & 0.8984  \\

 \hline
\end{tabular} 
\end{table}

\textbf{CNN:} Table \ref{tab:comp} shows that the recall is projected to be 0.9162 and the accuracy is assessed to be 92.38 percent for CNN. During the CNN training process, we achieved a validation accuracy of 92\%. The model was trained over the course of 40 epochs, with the validation accuracy measured at 98.80\% and the validation loss measured at 0.0303.The model was not developed further because overfitting was a possibility. The precision number, which can be seen as 0.9667, indicates false positive situations, but the recall value, which can be seen as 0.9162, reveals CNN's limitations in the case of false negative cases. Fig. \ref{fig:img2} shows the confusion matrix of CNN for the testing samples.
 \begin{figure}[!h]
    \centering
    \includegraphics[width=0.53\textwidth]{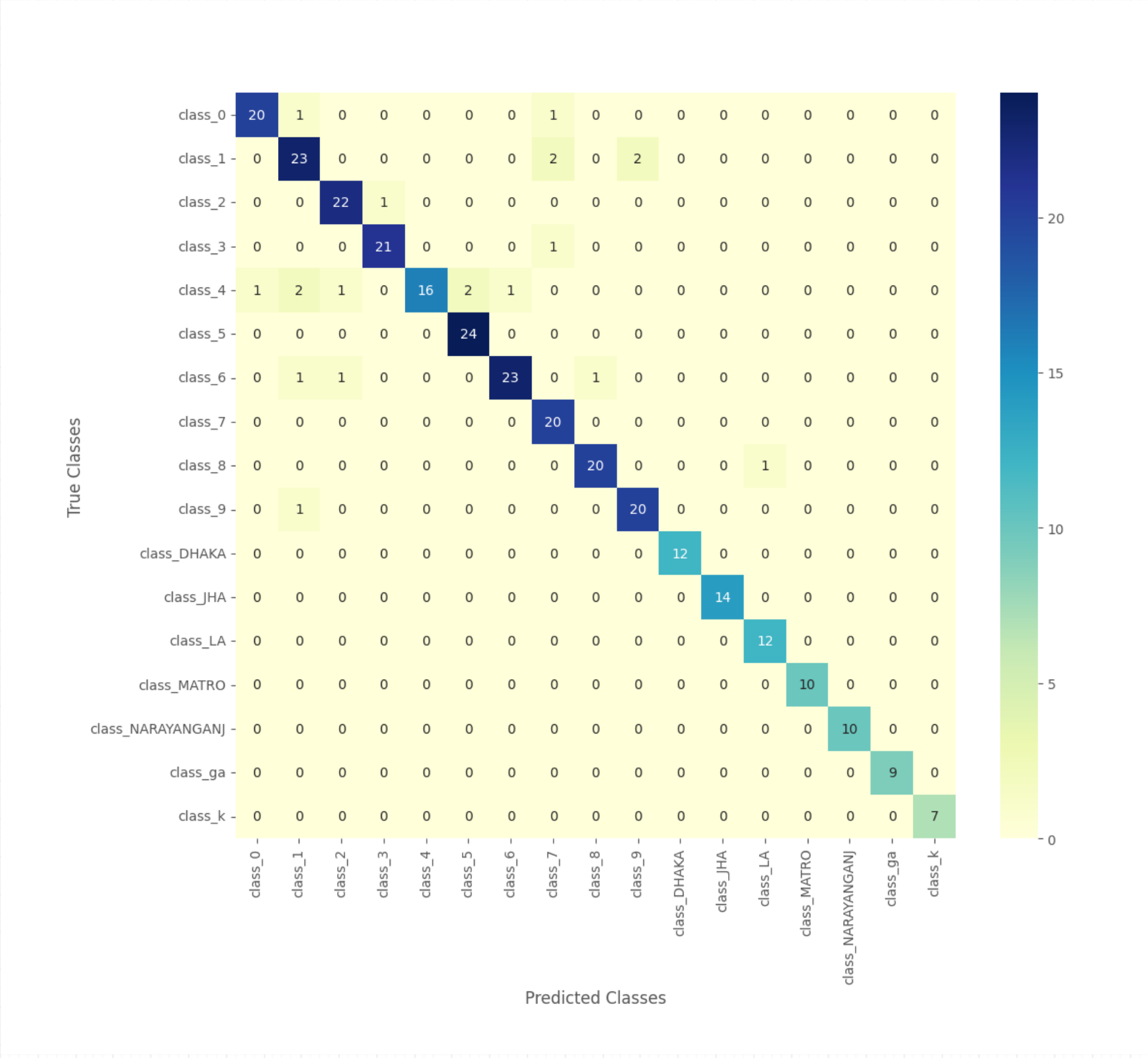}
    \caption{Confusion Matrix of the proposed CNN architecture.}
    \label{fig:img2}
\end{figure}

\textbf{ResNet50:} The ResNet50 model demonstrates strong overall performance with an accuracy of 87.50 \%, indicating its ability to correctly classify the majority of samples in the dataset. To train the model 30 epochs were used. The high precision value of 0.9343 signifies that the model is cautious in predicting positive instances, resulting in a low number of false positives. Moreover, the recall score of 0.8827 suggests that the model effectively captures a significant portion of the actual positive samples, reducing false negatives. The F1-Score of 0.8984 further validates the model's balanced performance between precision and recall. In summary, ResNet50 exhibits notable accuracy, precision, and recall, making it a reliable choice for the given classification task.

Nevertheless, there is a deficiency in ResNet50's ability to accurately detect Bengali characters on number plates. Figure \ref{fig:resnetfault} illustrates instances of incorrect predictions made by ResNet50. While the model is capable of accurately predicting Bengali numerals, it occasionally struggles to properly identify certain Bengali characters on the plates, specifically those used for city names and registration letters
\begin{figure}[!h]

  \centering
  \includegraphics[width=0.5\textwidth]{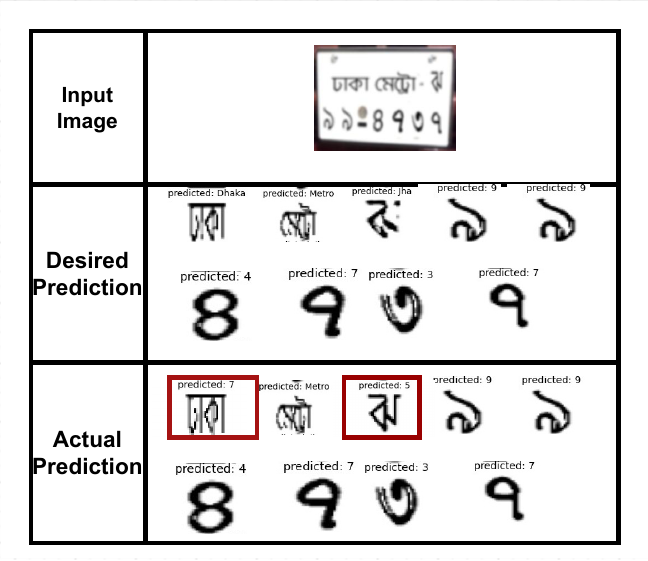}
   \caption{ Unveiling Model Challenges- Exploring an Incorrect Prediction by ResNet50}
   \label{fig:resnetfault}
\end{figure}

\textbf{Performance of our methodology:} We utilized the SequenceMatcher class of difflib library which compares pairs of sequences of any type to evaluate our 30 testing samples (number plates), achieving an 88\% accuracy on average.Some examples of SequenceMatcher performance are given in Table \ref{tab:scoresample}. While the CNN-based predictions for the plates were generally robust, occasional plate noise led to the inclusion of some distorted images in our model's output (Fig. \ref{fig:noise-image}). Consequently, slight alterations in the desired output sequences occurred. As a result, the overall accuracy was somewhat affected.
\begin{table}[h!]
      \centering
       \caption{ Sample Data - Desired vs. Generated Output with Corresponding Accuracy}
       \label{tab:scoresample}
\begin{tabular}{ c c c  }

 \hline
  \textbf{Desired Output}  & \textbf{Generated Output} & \textbf{Accuracy}  \\
 \hline
 9dhaka94matro773jha7 & 9Dhaka984Matro773Jha7 & 0.95  \\
 2dhaka2matro404ka4 & 2dhaka2matro404ka4 & 1  \\
 5dhaka47818jha & 553Dhaka47898Jha & 0.86  \\
 5dhaka2315ka6 & 5Dhaka231526 & 0.84  \\
\hline
\end{tabular} 
\end{table}
\subsection{ \textbf{Experimental Analysis}}
The accuracy of 88\% shows that there is a fair amount of similarity between the machine-generated translation and the reference translations. This accuracy is calculated by taking the average of all the 30 samples (number plate) that were taken to evaluate the model. Table \ref{tab:scoresample} shows some sample scores achieved by SequenceMatcher Library.   
Experimenting with various threshold values yielded diverse results. When employing a stringent 95\% threshold,
approximately 57\% (17 out of 30) of the number plates
achieved 95\% accuracy or higher. Conversely, a 90%
threshold led to over 80\% (24 out of 30) exceeding 90\%
accuracy, emphasizing broader accuracy without compromising quality

As we are detecting the Bengali digits and letters from poor-quality images by maximizing the image quality using several image processing techniques, there remain some excessive noises in the image (Fig. \ref{fig:noise-image}). If these noises could be minimized the model would have achieved a higher accuracy. 


\begin{figure}[!h]

  \centering
  \includegraphics[width=0.4\textwidth]{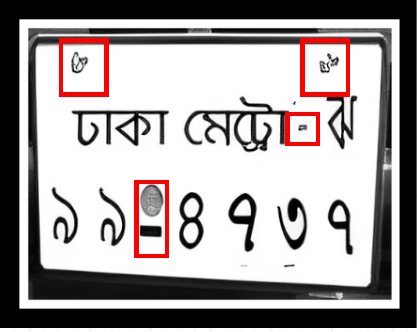}
   \caption{ Uncovering Image Noise: How Garbage-Induced Distortions Affect Model Accuracy}
    \label{fig:noise-image}
\end{figure}
We face some issues with the spacing of characters. In Bengali language ”Matra” is employed to connect the alphabet in a word (a line connects multiple characters). It is believed that the entire word ”Matra” consists
of one connected component. Sometimes the matra gets disconnected but it still represents one connected component. It made it hard for the model to recognize these components. The image segmentation part is also challenging while solving these problems.

\subsection{Performance Comparison}
Table \ref{tab:comparision-table} provides a brief comparison of various papers. Kabiraj et al. \cite{kabiraj2023number} uses increased super-resolution with ESRGAN and CNN for enhanced character recognition. Haque et al. \cite{haque2022automatic} uses ESRGAN-based image reconstruction to obtain more accuracy in reading Bengali license plates. With the use of cutting-edge methods like GFPGAN, morphological processing, CNN, and RestNet50, our suggested system demonstrates Automated License Plate Recognition (ALPR) with digit recognition, generating better results.

\begin{table}[]
\caption{Comparison of our proposed approach with others.}

\begin{tabular}{cccc}
\hline
\textbf{Authors}                                           & \textbf{Architecture} & {\begin{tabular}[c]{@{}c@{}}\textbf{Resolution}\\ \textbf{Enhance-}\\\textbf{ment}\end{tabular}}  & 
\textbf{Accuracy}                                                                                           \\ \hline
\multirow{2}{*}{Proposed}                                  & CNN                  & \multirow{2}{*}{\begin{tabular}[c]{@{}c@{}}GFPGAN \end{tabular}} & \begin{tabular}[c]{@{}c@{}}88\% \end{tabular} \\  & ResNet50              && \begin{tabular}[c]{@{}c@{}}83\% \end{tabular} \\ \hline
Kabiraj et al. \cite{kabiraj2023number} & CNN                 & ESRGAN                                                                                              & \begin{tabular}[c]{@{}c@{}}84\% \end{tabular}               \\ \hline
Haque et al. \cite{haque2022automatic}     & PSNR                  & ESRGAN                                                                                              & \begin{tabular}[c]{@{}c@{}}78\% \end{tabular}         \\ \hline
\end{tabular}
\label{tab:comparision-table}

\end{table}

\section{\textbf{CONCLUSION AND FUTURE WORK}}
This research study presents a method for identifying vehicle license plates in Bangladesh, written in Bengali. 
CNN provides a wealth of features to enhance character recognition accuracy. 
The system was evaluated using 1292 images of Bengali digits and characters and achieved an accuracy of 88\%. 
In terms of future scope, this system can be enhanced by adding other image processing methods such as noise reduction, perspective correction, and picture enhancement to further improve the precision of license plate recognition. To effectively control traffic and increase security, the system can also be connected with real-time surveillance systems. The system is adaptable for use around the world and can be expanded to recognize license plates from more nations and languages.

\bibliographystyle{ieeetr}
\bibliography{biblio}

\end{document}